\documentclass{llncs}
\usepackage{epsf}

\usepackage{makeidx}  
\usepackage{framed,multirow}
\usepackage{times}
\usepackage{latexsym}
\usepackage{graphicx}
\usepackage{amsmath}
\usepackage{amsfonts}
\usepackage{amssymb}
\usepackage{mathrsfs}
\usepackage[utf8]{inputenc}
\usepackage{color}
\usepackage{xcolor}
\usepackage{bbm}
\usepackage{url}
\usepackage{algorithm2e}
\usepackage{background}
\backgroundsetup{
  position=current page.east,
  angle=-90,
  nodeanchor=east,
  vshift=-5mm,
  opacity=0.5,
  scale=3,
  contents=Accepted draft MICAI-IJCLA
}

\begin{document}

\title{Low-resource bilingual lexicon extraction using graph based word embeddings}


\author{Ximena Gutierrez-Vasques\ \and Víctor Mijangos}

\institute{UNAM, Mexico City,\\
\email{xim@unam.unam.mx, vmijangosc@ciencias.unam.mx},\
}

%
%

\maketitle              

\begin{abstract}
In this work we focus on the task of automatically extracting bilingual lexicon for the language pair Spanish-Nahuatl. This is a low-resource setting where only a small amount of parallel corpus is available. 
Most of the downstream methods do not work well under low-resources conditions. This is specially true for the approaches that use vectorial representations like Word2Vec. Our proposal is to construct bilingual word vectors from a graph. This graph is generated using translation pairs obtained from an unsupervised word alignment method. 

We show that, in a low-resource setting, these type of vectors are successful in representing words in a bilingual semantic space. Moreover, when a linear transformation is applied to translate words from one language to another, our graph based representations considerably outperform the popular setting that uses Word2Vec.

\end{abstract}

\section{Introduction}

In natural language processing (NLP), bilingual lexicon extraction is the task of automatically obtaining pairs of words that are translations of each other. This has been an active area of research for several years, since it constitutes an important step in statistical machine translation systems. Moreover, it is helpful for building bilingual lexicons or dictionaries and for bilingual terminology extraction.
The availability of big amounts of parallel and monolingual corpora allow to model the relations between words of two different languages. There exists different approaches to estimate the lexical correspondences. For instance, through probabilistic translation models for word alignment or using association and similarity measures.

Lately, there has been a strong interest in vectorial word representations that capture meaning through contexts. This is a linguistics inspired notion, based on the distributional hypothesis \cite{harris1954distributional,firth1957synopsis}. In a multilingual setting, the idea is that a word that occurs in a given context in a language should have a translation that occurs in a similar context in the other language.
If we have vectors representing lexical units of two different languages, we can project them into the same space and compute distances in order to find translation candidates.

However, in order to be able to build this type of  representations, a huge amount of data is needed. In fact, most of the methods to perform bilingual lexicon extraction require big parallel or monolingual corpora.
Many languages of the world do not have big amounts of readily available corpora, specially low-resource languages. Due to this, a current research question in NLP is how to make downstream methods to work properly in low-resource settings.

We take as case of study the language pair Spanish-Nahuatl. These languages are spoken in the same country (Mexico) but they belong to different linguistic families. In the case of Nahuatl, there is scarcity of monolingual and parallel documents.
We face a low-resource setting where popular methods for bilingual lexicon extraction do not work properly. We propose the combination of different approaches to improve the estimation of the bilingual correspondences, based on a small parallel corpus.
In particular, we create a graph taking into account association scores between words of the two different languages. From this graph we generate vectors that are more suitable for dealing with our low-resource setting.

Our approach tries to be as unsupervised as possible, since it is difficult to rely on Nahuatl resources due to its scarcity and lack of standardization. By performing bilingual lexicon extraction, we would like to contribute to the machine-readable resources for this language pair, since bilingual dictionaries are expensive resources that are not always available for all
language pairs. Specially, when one of the languages is resource-poor.

The structure of the paper is as follows: Section~\ref{sec:back} shows a general description of some main concepts that we use for the development of our work. Section~\ref{sec:metod} contains the proposed methodology to extract bilingual lexicon and generating graph based embeddings. In section~\ref{sec:results} we describe the evaluation and the obtained results. Finally, Section~\ref{sec:conc} contains the conclusions and future work.  

\section{Background} \label{sec:back}
\subsection{Bilingual lexicon extraction}


\subsubsection{Estimation and association approaches for bilingual lexicon extraction}
Many methods of bilingual lexicon extraction assume the availability of big amounts of parallel corpora that allow to model the relations between lexical units of the translated texts. This is fundamental in terms of statistical machine translation (SMT), where word and phrase alignments are necessary in order to translate a whole sentence.

From the traditional SMT perspective there has been two main approaches for finding the bilingual correspondences of words: estimation and association.

In the estimation approach, a probabilistic distribution is estimated through a likelihood maximization process. Some of the most popular methods for word alignment are the IBM-models \cite{brown1993mathematics}, they are based in a estimation approach, they produce probabilistic tables of lexical translations \cite{brown1993mathematics}.

On the other hand, association based approaches (also known as heuristic methods) rely on similarity or association measures. They calculate a score that determines which pairs of words are suitable translations. There are many works that use association measures, e.g., log-likelihood measures \cite{tufics2002revealing}, t-scores \cite{ahrenberg1998simple}, positional difference between two successive occurrences of a word \cite{fung2000statistical}, sub-sampling of a parallel corpus \cite{lardilleux2009sampling} just to mention some.

Since the quality of word alignment methods is heavily dependant on the amount of parallel data. There are alternative approaches, e.g., some works assume that if there is not enough parallel corpora for a language pair, there is enough comparable corpora or monolingual corpora for each of the languages. In these approaches bilingual lexicons are induced by taking into account several features, e.g, contextual features \cite{firth1957synopsis,harris1954distributional} as described below.

\subsubsection{Linear mapping between languages} \label{secc:mikolov}
Recent works use distributional and distributed vector representations (word embeddings). These vectorial representations based on contexts of the words, have became useful to find translations between different languages.The idea is to build multilingual representations or to map monolingual vectors to the same space in order to find the closest translations \cite{lauly2014learning,hermann2014multilingual,mikolov2013exploiting}. These state of the art methods may not require parallel corpora but they are still based on huge amounts of monolingual or comparable corpora in order to work properly. It has been shown that when they face a low resource setting they can have even worst performance than less sophisticated methods \cite{levy2016strong}.

From these works, we will deepen into the ones focused on linear mapping between language representations in order to find bilingual correspondences. \cite{mikolov2013exploiting} proposed a way to find a linear transformation that maps vectors from one language to another . In the standard setting, it is necessary to induce separately two monolingual embedding spaces, $\mathbb{R}^{d}$ and $\mathbb{R}^{d_2}$. This can be done by using big amounts of monolingual corpus for each language. Word embedding models such as CBOW or Skip-Gram can be used \cite{mikolov2013efficient}.
In order to learn a linear mapping between the two spaces, we need a supervised signal. This supervised signal is a seed lexicon that contains a list of translation pairs $(x_i, y_i)$, where $x_i \in \mathbb{R}^{d}$, $y_i \in \mathbb{R}^{d_2}$

The idea is to exploit geometrical properties of the embedding vector spaces. \cite{mikolov2013exploiting} assumes that, when the corpora is comparable, certain geometric regularities remain in the different languages spaces. These regularities are supposed to be linear. Then the goal is to find a linear transformation $T: \mathbb{R}^d \to \mathbb{R}^{d_2}$; this is, the transformation take a word vector form the first language and transform it into the correspondence translation in the second language. If $x_i \in \mathbb{R}^d, i=1,2,...,N$ are a word vectors in the source language and $y_i \in \mathbb{R}^{d_2}, i=1,2,...,N$ are the word vectors of the possible translations, then $T$ can be learned by solving the next optimization problem:

\begin{equation}
    \min_T \sum_{i=1}^N ||T(x_i) - y_i||^2 + \lambda ||T||^2
\end{equation}

The right hand side of the sum is a parametrization term with $\lambda \in \mathbb{R}$. This optimization problem defines a least square problem which can be solved by a stochastic gradient descent method.

\subsection{The Node2Vec architecture}

Node2vec is an algorithm for learning vectorial continuous representations based on a graph \cite{grover2016node2vec}.
It is similar to the Word2Vec formulation \cite{mikolov2013efficient}. Nevertheless, Node2Vec takes as input a graph structure $G=(V,E,\phi)$ where $V$ is the set of nodes, $E$ is the set of edges and $\phi: E\to \mathbb{R}$ is a weight function over the edges. It is not necessary for the graph to be weighted. However, for the purpose of this work, we consider a weighted graph.

The aim of the Node2Vec algorithm is to find a mapping $F: V \to \mathbb{R}^d$ transforming the nodes of a graph into vectors of a $d$-dimensional vector space. Here, just like in Word2Vec, $d$ is a parameter that specifies the dimensionality of the output vectors. 
Node2Vec formulation is a maximum likelihood risk minimization problem, that in general can be defined as the following functional \cite{vapnik1998statistical}:

\begin{equation} \label{ML}
    Q(x,\theta) = -\log q(x|\theta)
\end{equation}

Where $x$ represents the data and $\theta$ the learning parameters. Node2Vec takes a neighborhood $N_S(v) \subseteq V$ of a node $v\in V$ by choosing a sampling strategy $S$ \cite{adamic2003friends}. Then, if $f(v) \in \mathbb{R}^d$ is the vector representation a the node $v\in V$, the maximum likelihood problem in Eq.~\ref{ML} takes the form:

\begin{equation}
    Q(v,f(v)) = -\log p(N_S(v) | f(v))
\end{equation}

where $p$ is the a conditional probability density function. And the empirical risk functional is defined as follows:

\begin{equation}
    R_{emp}(f(v)) = - \frac{1}{n} \sum_{v\in V} \log p(N_S(v) | f(v)) 
\end{equation}

For \cite{grover2016node2vec} this empirical risk functional is minimizing by maximizing the dual problem. Then, the objective function can be established as:

\begin{equation}\label{obj}
    \max_f \sum_{v \in V} \log p(N_S(v) | f(v)) 
\end{equation}

In order to make tractable this optimization problem, the authors assumed conditional independence on the distribution. This way, the conditional probability can be defined as: $$p(N_S(v) | f(v)) = \prod_{u\in N_S(v)} p(u|f(v))$$ Also symmetry is assumed to compute a symmetric neighbors probability \cite{hinton2003stochastic} or Softmax function:

\begin{equation}
    p(u|f(v)) := \frac{ exp\{ \langle f(u), f(v) \rangle  \} }{ \sum_{u'\in V} exp\{ \langle f(u'), f(v) \rangle  \} }
\end{equation}

This simplifies the objective function of Eq.~\ref{obj} to:

\begin{equation}\label{obj2}
\max_f \sum_{v\in V} [\sum_{u' \in N_S(v)} exp\{ \langle f(u'), f(v) \rangle \} -\log \sum_{u \in V} exp\{ \langle f(u), f(v) \rangle  \} ]
\end{equation}

Eq.~\ref{obj2} is similar to the optimization problem proposed for word embeddings \cite{mikolov2013efficient} and it can be solved with negative sampling \cite{goldberg2014word2vec}. In particular, the Node2Vec algorithm implemented by \cite{grover2016node2vec} uses for sampling strategy, $S$, the Alias method \cite{walker1974new,kronmal1979alias}. 

\section{Methodology} \label{sec:metod}

\subsection{The corpus}
We focus on the language pair Spanish-Nahuatl. These two languages are spoken in the same
country but they are distant from each other, i.e., different syntactic and morphological phenomena. Making it harder to find bilingual correspondences.
Nahuatl is an indigenous language with around 1.5 M speakers and it is a language with scarcity of monolingual and parallel corpora. Most of the documents that can be easily found in Nahuatl are translations from (or to) Spanish, therefore it is easier to obtain parallel corpora than monolingual one.

\begin{table}[h]
    \begin{center}
    \begin{tabular}{|c|c|c|c|}
        \hline \bf Language & \bf Tokens & \bf Types& \bf Sentences \\ \hline
        Spanish (ES) & 118364 & 13233 &5852 \\ \hline
        Nahuatl (NA) & 81850 & 21207 &5852\\ \hline
    \end{tabular}
    \end{center}
    \caption{\label{parallelcorpus} Size of the parallel corpus }
\end{table}

We are dealing with a low-resource setting, since the amount of available parallel corpora is small in terms of the amount of text required by popular NLP models. We used a parallel corpus created for these languages by \cite{gutierrez2016axolotl}. This corpus was mainly extracted from non digital books. It contains documents from different domains and it has a lot of dialectal and orthographic variation. 
The lack of orthographic normalization is an issue that has a negative impact in NLP tasks, since there can be different word forms associated to the same word. In order to counteract this variation, we only used a subset of the parallel corpus that had, more or less, the same orthographic norm.  Table \ref{parallelcorpus} shows the size of the parallel corpus that we used for this work. The corpus is aligned at the sentence or paragraph level.


\subsection{Overall method}
The overall procedure can be summarized into the following algorithmic steps. The next subsections contain the detailed explanation of each stage.

\begin{enumerate}
\item For each Spanish word, a scored list of translation candidates is extracted using a sampling-based method
\item Once the translation scores are obtained, a graph is computed. The nodes correspond to the vocabulary words of each language. The weighted connections between the nodes are obtained from the scored lists from previous step
\item Word vectors are computed from this graph, i.e., using Node2Vec each node (a word) is transformed to a continuous vector in $\mathbb{R}^n$
\item A linear transformation is learned in order to map from Spanish vector space to Nahuatl vector space. A seed lexicon of correct translation pairs is used to learn this transformation  
\item Once the transformation is calculated, it is applied to a set of evaluation Spanish words. For each projected vector, the nearest vectors correspond to the translation candidates in Nahuatl ($L_2$ metric was used).
 
\end{enumerate}

\subsection{Sampling based bilingual extraction}
The first stage of our methodology implies the extraction of translation word pairs, using an association based method. In particular, we used a method called sampling-based or Anymalign \cite{lardilleux2009sampling}, where
only those words that appear exactly in the same sentences are considered for alignment. The idea of this method is to produce more candidates by creating many sub-corpora of small sizes (sub-sampling).
We chose this method because it has shown better results compared to estimation approaches, for one to one translations \cite{luo2013comparison}. 

It is important to mention that in order to improve the performance of this method applied to our small parallel corpus, morphology of the languages must be taken into account. We used lemmas for Spanish and
morphs for Nahuatl. \cite{ximena2017bilingual} shows that using this morphological text representations, achieved the best performance in the task of extracting bilingual lexicon.
These extracted translation pairs will be later used by Node2Vec and for automatically creating a bilingual seed lexicon.

\subsection{Graph based embeddings}
We have mentioned before, that vectorial representations, e.g.,  Word2Vec, have became popular in NLP. These real valued, non sparse representations encode the meaning of a word based on its contexts.

Unfortunately, we are dealing with a low-resource setting where only a small amount of texts are available. Word embedding methods needs a big number of data to reach a suitable probabilistic distribution in the vector space \cite{bengio2003neural,mikolov2013efficient}. 
Training this type of embeddings with small amounts of text leads to no good representations. Therefore, multilingual tasks like the one described in Section~\ref{secc:mikolov}, do not achieve good results. 

In order to solve this issue, we propose an intermediate layer in the construction of the embeddings. The sample-based method that we used in the previous section, returns for each word a list of translation candidates ranked by a score. This can be seen as a graph structure. We can take advantage of this graph structure to perform bilingual embeddings \cite{yan2014graph}.

Let $C_1$ and $C_2$ be two parallel corpora in two different languages. And let $L_1$ and $L_2$ their respective lexicons (sets of words of each corpus). We construct a graph related $L_1$ and $L_2$. If $w^1 \in L_1$ and $w^2 \in L_2$, then $(w^1,w^2) \in E$ if $w^2$ is a possible translation to $w^1$, according to the translation pairs obtained by the sampling-based method. The weight $\phi(w^1, w^2)$ is obtained by the scores of the sampling-based method explained earlier.

It is important to mention some of the graph parameters. We selected 5 as the number of random walks, 100 iterations in the learning procedure and 500 dimensions for the output vectors. The resulting graph had 5474 nodes, it was undirected, i.e., the relations between the words and its translations were considered as symmetric. 

Once we obtained the graph $G = (V,E, \phi)$, we generate the vector representation for each node by applying Node2Vec. To generate the vectors, we used the default parameters in Node2Vec, except that we took the undirected weighted graph option. The output vectors were of dimension 128.
The idea is to obtain vectors representations $f(w^1), f(w^2) \in \mathbb{R}^d$ such that if $w^2 \in N_S(w^1) \subseteq V$ then $f(w^2) \in N_{||\cdot||}(f(w^1)) \subseteq \mathbb{R}^d$. Here $||\cdot||$ represents a metric over $\mathbb{R}^d$ determining neighborhoods or others similarity measures \cite{corley2005measuring,pedersen2007measures}. Given the methodology for constructing the graph, the neighbors for $f(w^1)$ in the vector space $\mathbb{R}^d$ are possible translations. The objective function describes the probability of a given neighborhood to be a set containing translation candidates for $w^1 \in L_1$. 

\begin{figure}
    \centering
    \includegraphics[scale=0.26]{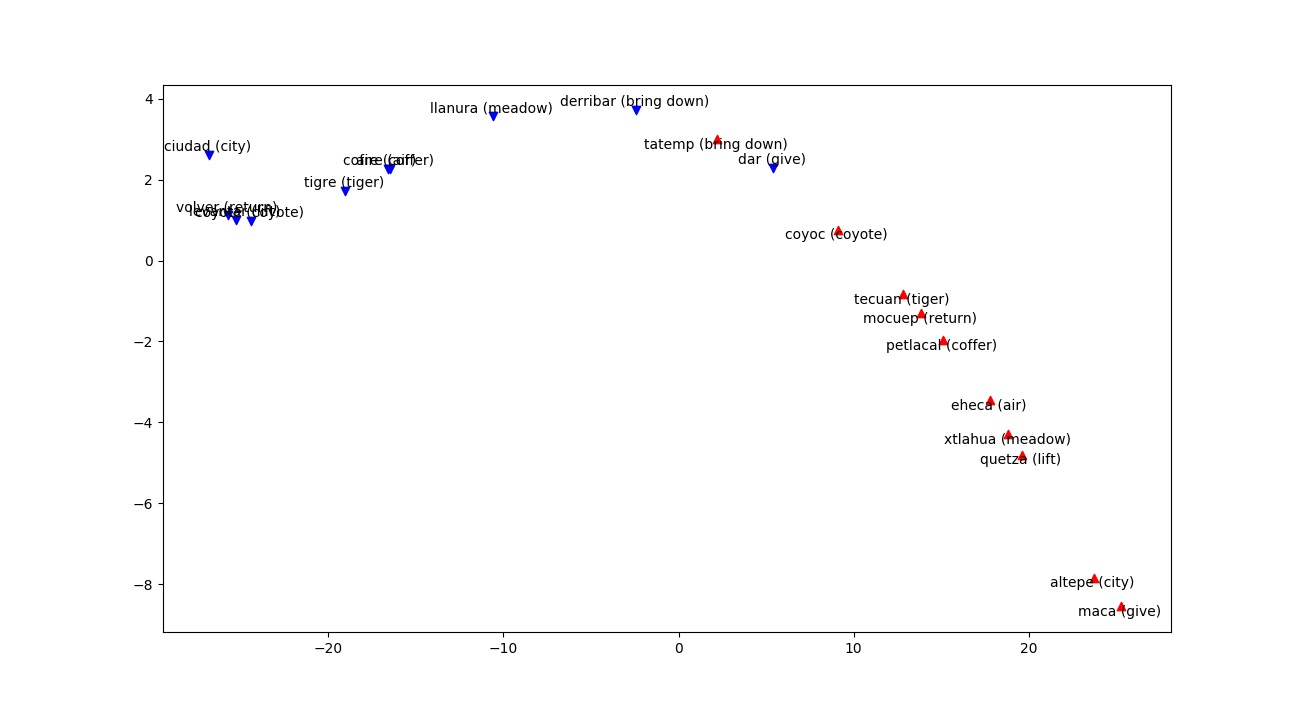}
    \caption{Word vectors using Word2Vec}
    \label{fig:w2v}
\end{figure}

\begin{figure}
    \centering
    \includegraphics[scale=0.26]{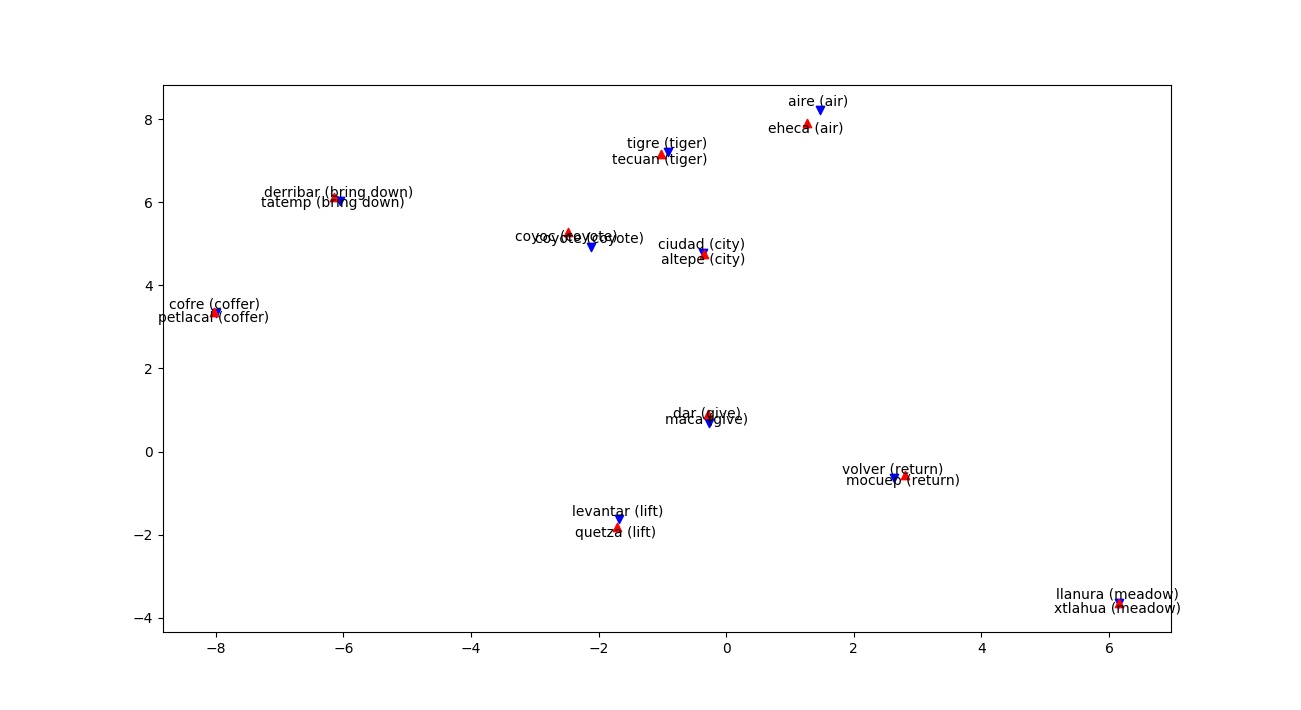}
    \caption{Word vectors using Node2Vec}
    \label{fig:n2v}
\end{figure}

Figure \ref{fig:w2v} shows the plot of some words using Word2Vec (CBOW, 300 dimensions). Figure~\ref{fig:n2v} shows the same words but using our graph based representations. We used t-SNE technique \cite{hinton2003stochastic,maaten2008visualizing} to plot the data.

\subsection{Linear mapping using graph based embeddings}

In order to learn a transformation matrix that maps between the two semantic spaces, we need a seed lexicon. Most of the works, extract this set of words from a bilingual dictionary. However, we wanted to keep our approach as unsupervised as possible. We did not use a pre-compiled seed lexicon, since it is difficult to rely in a single standardized dictionary for Nahuatl.

We created our seed lexicon by combining two word alignment methods: IBM-Model 1 \cite{brown1993mathematics} and sampling based (Anymalign) \cite{lardilleux2009sampling}. We combine them in the following way: 1) For each method we extract the translations pairs that are symmetric, i.e., word pairs that obtain the same translation in both ways (Spanish to Nahuatl and Nahuatl to Spanish); 2) we make a ranking of translations pairs, in the top we put the symmetric pairs that are equal in both methods, after these we put the rest of symmetric pairs obtained with either of the two methods; 3) this is our seed lexicon, we can make an arbitrary cut to filter the first $n$ entries.
The idea is that translation candidates that are symmetric, and that match using two different approaches, could be highly reliable, therefore suitable for the seed lexicon.

Our seed lexicon contains 553 entries. Afterwards, we calculate the transformation matrix that maps Spanish words to Nahuatl space, as explained in Section~\ref{secc:mikolov}. We learned the transformation using two different types for embeddings: Word2Vec and our formulation of bilingual Node2Vec.

The linear map $\hat{W}$ was defined by an objective function as follows: $$\hat{W} = \arg\min_W ||W w^1 - w^2||_2^2 + \gamma ||W||_2^2$$ Where $w^1$ is a word vector representation in Spanish and $w^2$ is the corresponding word vector in Nahuatl.
The main idea of the linear mapping is to exploit the geometric characteristics of both languages representations in vector space to get closer the corresponding translation to the original word.

In the next section, we detail the obtained results.

\section{Analysis of results} \label{sec:results}

First, it is important to notice the obtained plots using the different types of embeddings. Using Word2Vec led to representations of words that are not close to their correspondent translation. It will be probably hard for the linear mapping approach to find a geometrical arrangement that relates both languages (Figure~\ref{fig:w2v}).

On the other hand, the representations obtained with our bilingual Node2Vec formulation seem more coherent. Word vectors appear near to their translation, this is expected since the graph was constructed using the weights obtained from a word alignment method. However, we conjecture that the graph structure captures relations that are not evident for a weighted procedure. Furthermore, association or estimation approaches to bilingual lexicon extraction do not build vectorial representations of words. We have created vector space representations, which would allow to perform operations and other procedures that requires vectors at input data \cite{baroni2014frege}.

In order to evaluate the representations in a bilingual lexicon extraction task, we constructed an evaluation set by randomly selecting 130 Spanish words with frequency greater than 2 in the corpus. Human annotators wrote possible translations for this set of words. It is important to notice that the seed lexicon used for the linear mapping did not contain any of these words to avoid overfitting. 

We used precision at one, precision at five and precision at ten, i.e., take into account up to 10 closest word vectors to a source word that we want to translate. 
First, we evaluated our graph based representations without using any linear mapping (\emph{N2V-NOmap}). Then, we applied linear mapping to the Node2Vec representations (\emph{N2V-map}) and the Word2Vec ones (\emph{W2V-map}). Results are shown in Table \ref{tab:results}. 
\begin{table}[!h]
    \centering
    \begin{tabular}{|c|c|c|c|} \hline
        ~ & \textbf{P@1} & \textbf{P@5} & \textbf{P@10} \\ \hline
        N2V-NOmap & 0.260 & 0.598 & 0.661 \\ \hline
        N2V-map & \textbf{0.614} & \textbf{0.835} & \textbf{0.866} \\ \hline
        W2V-map & 0.102 & 0.125 &0.164
        \\ \hline
    \end{tabular}
    \caption{Evaluation of bilingual lexicon extraction using different vector representations }
    \label{tab:results}
\end{table}

The bilingual graph based embeddings, clearly outperform the popular setting of mapping Word2Vec representations between languages \cite{mikolov2013exploiting}. This is consistent with recent works that have pointed out that the latter approach, tends to have a bad performance, specially if they lack of huge amounts of corpora to train the vector space representations \cite{levy2016strong}.
In this sense, our proposal seems to be successful in dealing with low-resource languages. We only needed a small parallel corpus (sentence aligned) and we were able to generate word vector representations that are useful for bilingual lexicon extraction.

It is interesting to notice that the graph based representations by themselves (\emph{N2V-NOmap}) are useful to detect correct translations. However, they are even more accurate if a linear transformation is applied (\emph{N2V-map}).

Although this is not the first work that combines graph structures and vector representations, our contribution lies in the fact that it is suitable for our low-resource setting (small parallel corpus). Similar methods \cite{pelevina2017making,newman2017second,blunsom2014multilingual,kovcisky2014learning} are only tested in rich resource settings where considerable bigger amounts of corpora are used to build representations. Their performance tends to drop under low-resource settings. Our approach improves the vector representations by having a prior distribution based on unsupervised bilingual extraction, combined with a graph representation.

\section{Conclusions} \label{sec:conc}
In this work we have presented an architecture to build multilingual vector representations, using a graph structure and word translation scores obtained from an associative sampling based method. 
Our main interest was to be able to perform bilingual lexicon extraction. We deal with a low-resource setting where only a small parallel corpus is available, it is not easy to find digital resources for one of the languages and the two languages are very distant from each other.

Our formulation seems to be successful in representing the words in a bilingual vector space. This is useful since, when working with low-resource languages, typical word embeddings are not of good quality due to the lack of training data.
Moreover, when we applied a linear mapping between the two languages, we were able to outperform the results obtained with Word2Vec representations.

With this work, we would like to contribute to the development of automatic translation technologies for Spanish-Nahuatl, since there are practically no technologies developed for indigenous languages of Mexico.
As future work, we would like to explore the operations like composition that can be applied to the obtained word vectors in order to represent phrases or larger units. Additionally, more complex weighting schemes can be proposed to generate the bilingual graphs and the graph based vectors.

\bibliographystyle{splncs}
\bibliography{biblio}

\end{document}